\documentclass{article}
\usepackage[final]{nips_2017}

\usepackage{float}
\usepackage[utf8]{inputenc} % allow utf-8 input
\usepackage[T1]{fontenc}    % use 8-bit T1 fonts
\usepackage{hyperref}       % hyperlinks
\usepackage{url}            % simple URL typesetting
\usepackage{booktabs}       % professional-quality tables
\usepackage{amsfonts}       % blackboard math symbols
\usepackage{nicefrac}       % compact symbols for 1/2, etc.
\usepackage{microtype}      % microtypography
\usepackage{bm}             % bold in math
\usepackage{graphicx}       % images
\usepackage{algorithm}      % algorithm
\usepackage[noend]{algpseudocode} % algorithm
\usepackage{caption}        % captionof
\usepackage{array}          % thick column hline
\usepackage{booktabs}       % table style
\usepackage{pbox}           % table line break
\usepackage{subcaption}     % multiple figures
\usepackage{amsmath}
\usepackage{textcomp,gensymb}

\title{Pushing the Limits of Capsule Networks}

\hypersetup{
    colorlinks = true,
}
\makeatletter
\def\BState{\State\hskip-\ALG@thistlm}
\makeatother
\floatname{algorithm}{Procedure}

\newcolumntype{?}{!{\vrule width 3pt}}

\author{
  Prem Nair, Rohan Doshi, Stefan Keselj \\
  Princeton University\\
  \texttt{\{pnair, rkdoshi, skeselj\}@princeton.edu} \\
  \\
  \textbf{December 2017}
}

\begin{document}

\maketitle

% \begin{abstract}

% \end{abstract}

\section{Motivation and Goal}

Convolutional neural networks use pooling and other downscaling operations to maintain translational invariance for detection of features, but in their architecture they do not explicitly maintain a representation of the locations of the features relative to each other. This means they do not represent two instances of the same object in different orientations the same way, like humans do, and so training them often requires extensive data augmentation and exceedingly deep networks.

A team at Google Brain recently made news with an attempt to fix this problem: Capsule Networks (\cite{sabour_dynamic_2017}), hereon referred to as CapsNets. While a normal CNN works with scalar outputs representing feature presence, a CapsNet works with vector outputs representing entity presence. There is much discussion about whether these new models could actually work as intended because \cite{sabour_dynamic_2017} mainly applied them to the MNIST dataset under favorable hyperparameter conditions.

We want to stress test CapsNet in various incremental ways to better understand their performance and expressiveness. In broad terms, the goals of our investigation are to:
\begin{enumerate}
\item Test CapsNets on datasets that are like MNIST but harder in a specific way.
\item Explore the internal embedding space and sources of error for CapsNets.
\end{enumerate}

\section{Background and Related Work}

The original idea driving this work is the capsule, first discussed in \cite{hinton_transforming_2011}. The capsule was pitched as a convenient way to represent an entity: it is a vector whose norm indicates the probability that entity is present and whose direction indicates the configuration that entity is in. Capsules could theoretically be combined to form hierarchical tree structures representing entities. It was a nice idea, but it did not get much traction until a few months ago when Sabour, Frosst, and Hinton finally figured out how to train a network to recognize and work in this space (\cite{sabour_dynamic_2017}). In this section we will attempt to convey our understanding of their model, but the best resource is their actual paper.

\subsection{Fundamentals of Capsules and the Dynamic Routing Algorithm }

CapsNets are comprised of layers of capsules, each of which are composed of neurons. But unlike neurons, capsules deal with inputs and an output that are vectors.  Each vector is meant to encode a rich representation of some entity: the vector's direction indicates what form the entity seems to be in (e.g. the pose of an object) and the vector's norm indicates the confidence of the representation. These vectors can be combined by affinely transforming and then adding them (analogous to multiplying scalars by weights and then adding them). The method for the forward pass of information from capsule layer to capsule layer is called dynamic routing.

A given capsule $j$  processes total input vector ${\bf s}_j$ and outputs an activation vector ${\bf v}_j$, whose direction is preserved, but whose magnitude is ``squashed'' between 0 and 1 by the non-linearity in Equation~\ref{squash}. The length of ${\bf v}_j$ indicates the probability of existence for the entity represented by the capsule.

\begin{equation}
{\bf v}_j = \frac{||{\bf s}_j||^2}{1+||{\bf s}_j||^2} \frac{{\bf s}_j}{||{\bf s}_j||}
\label{squash}
\end{equation}

The vector ${\bf s}_j$ is generated by processing the vectors from each of the connected capsules $i$ from the previous layers. For a given capsule activation from the previous layer $u_i$, we apply weight matrix $W_{ij}$ to yield $\bf\hat{u}_{j|i}$, a ``prediction'' vector to approximate ${\bf v}_j$. The $\bf\hat{u}_{j|i}$ from each capsule $i$ is weighted by a coupling coefficient $c_{ij}$, which is calculated via the dynamic routing algorithm.

\begin{equation}
{\bf \hat{u}}_{j|i} = {\bf W}_{ij}{\bf u}_i \ , \ \ \ \ \ \ \
{\bf s}_j = \sum_i c_{ij} {\bf \hat{u}}_{j|i}
\end{equation}

The sum of the coupling coefficients between capsule $i$ and all possible capsules $j$ in the next layer equals 1, forcing capsules to weigh the importance of more up-stream capsules. This is enforced by calculating $c_{ij}$ through a ``routing softmax'', whose inputs $b_{ij}$ are the log prior probabilities that capsule $i$ and $j$ are coupled. 

\begin{equation}
c_{ij} = \frac{\exp(b_{ij})}{\sum_k \exp(b_{ik})}
\label{softmax}
\end{equation}

\begin{figure}
 	\centering
	\includegraphics[width=\textwidth]{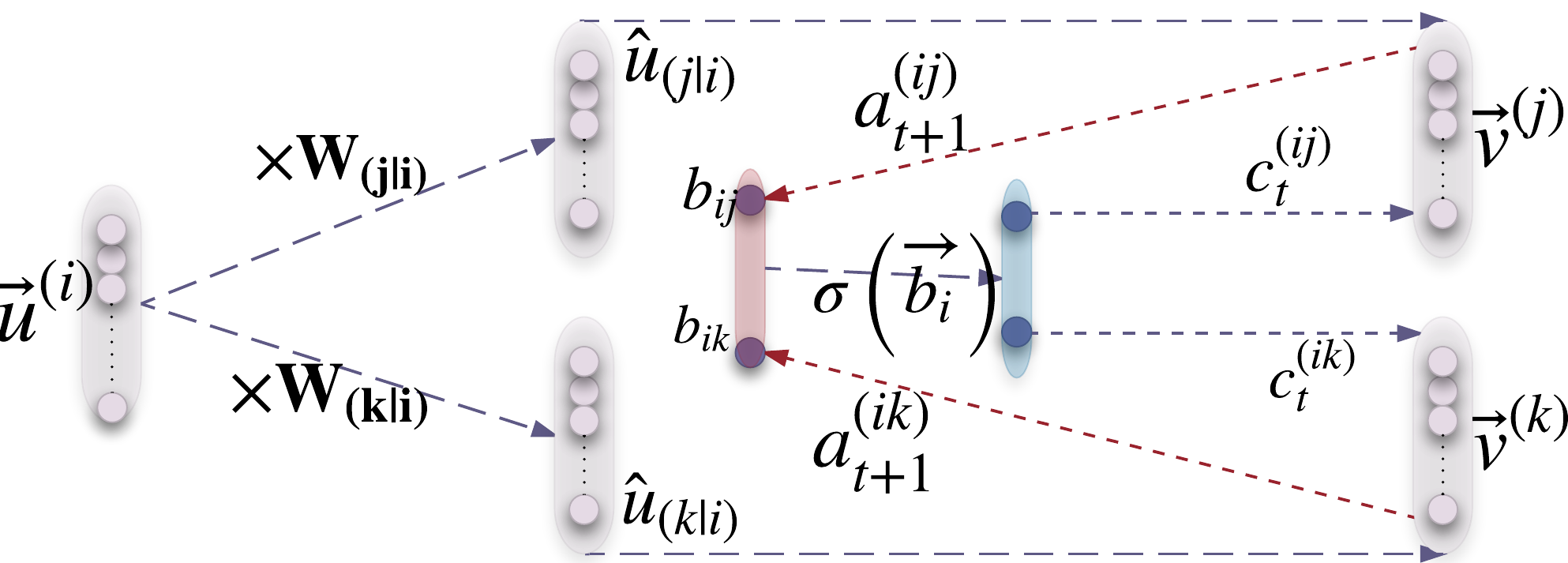}
	\caption{Capsule Routing (\cite{sabour_dynamic_2017})}
 	\label{capsRouting}
\end{figure}

Now, to update $c_{ij}$ via dynamic routing, we rely on a simple heuristic: we want to reward coupling coefficients that maximize the agreement $a_{ij}$ between the prediction and activation vector. In other words, we want to maximize $a_{ij}=u_{i|j} \cdot {\bf v}_j$ In order to update the coupling coefficients, $b_{ij}$ is iteratively incremented by $a_{ij}$ through iterations of the dynamic routing algorithm (Procedure~\ref{routingalg}) between training iterations for the model, and also visualized in Figure~\ref{capsRouting}.

\begin{algorithm}
\caption{Routing algorithm (from \cite{sabour_dynamic_2017})}\label{routingalg}
\begin{algorithmic}[1]
\Procedure{Routing}{$\bm{\hat{u}}_{j|i}$, $r$, $l$}
\State for all capsule $i$ in layer $l$ and capsule $j$ in layer $(l+1)$: $b_{ij} \gets 0$.
\For{$r$ iterations}
\State for all capsule $i$ in layer $l$: ${\bf c}_i \gets \texttt{softmax}({\bf b}_i)$ \Comment{\texttt{softmax} computes Eq.~\ref{softmax}}
\State for all capsule $j$ in layer $(l+1)$: ${\bf s}_j \gets \sum_i{c_{ij}{\bf \hat u}_{j|i}}$
\State for all capsule $j$ in layer $(l+1)$: ${\bf v}_j \gets \texttt{squash}({\bf s}_j)$ \Comment{\texttt{squash} computes Eq.~\ref{squash}}
\State for all capsule $i$ in layer $l$ and capsule $j$ in layer $(l+1)$: $b_{ij} \gets b_{ij} + {\bf \hat{u}}_{j|i} . {\bf v}_j$
\EndFor
\Return ${\bf v}_j$
\EndProcedure
\end{algorithmic}
\end{algorithm}

\subsection{CapsNet Architecture}

\begin{figure}
 	\centering
	\includegraphics[width=\textwidth]{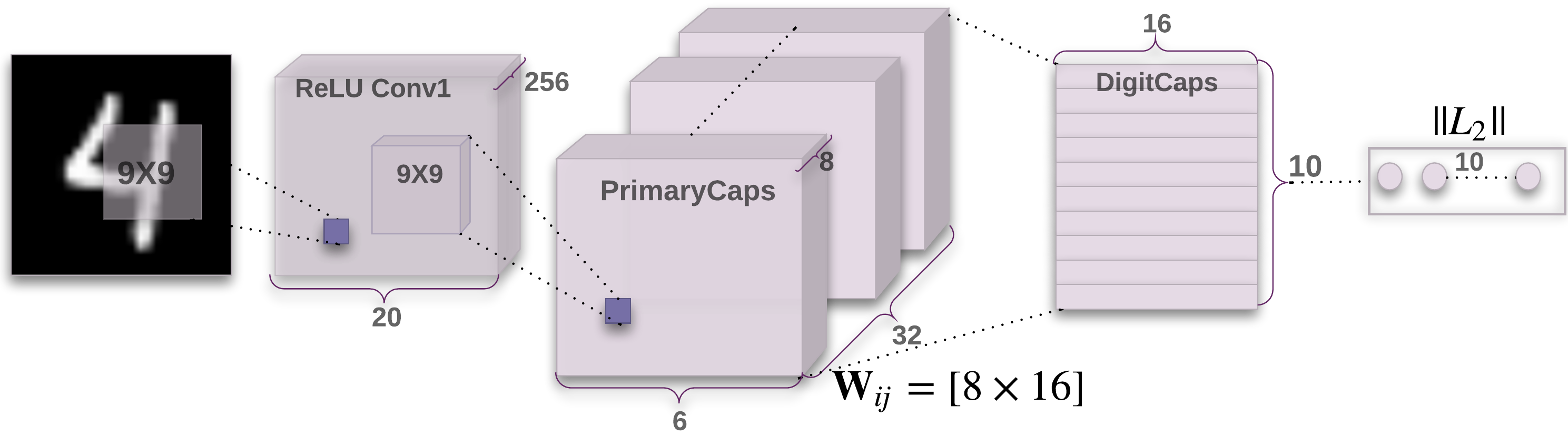}
	\caption{CapsNet Architecture (\cite{sabour_dynamic_2017})}
 	\label{capsnetArch}
\end{figure}

\begin{figure}
  	\centering
	\includegraphics[height=3cm]{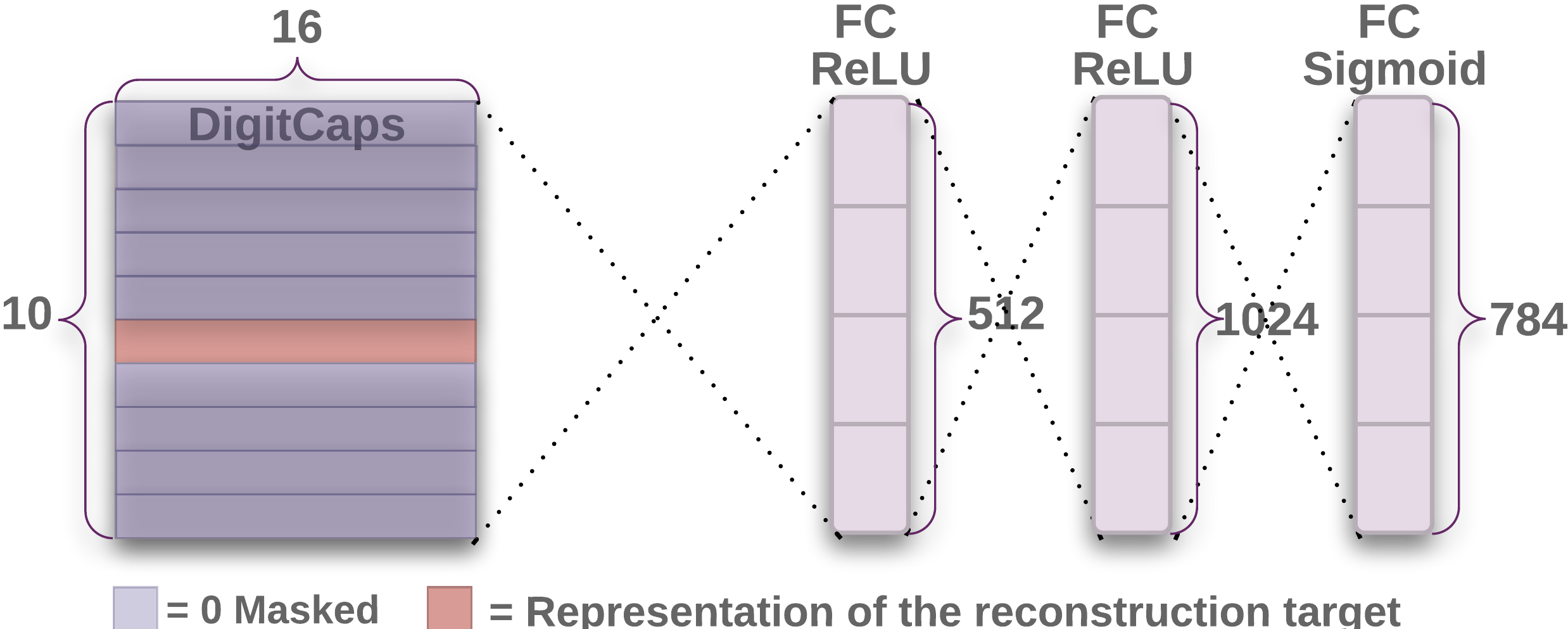}
	\caption{Reconstruction Architecture (\cite{sabour_dynamic_2017})}
 	\label{reconsArch}
\end{figure}

We work with a simple CapsNet architecture proposed in the original paper. It contains two halves. The first half (Figure~\ref{capsnetArch}) outputs a class prediction after three layers of processing (two convolution layers and one fully connected layer). The first layer is a convolutional layer that maps the $28\times28$ image to a $6\times6\times256$ volume using a $9\times9$ kernel, 256 feature maps, a stride of 1, and ReLU non-linearity. Next, the second layer (PrimaryCaps), is a convolution capsule layer that produces a $6\times6\times256$ volume using a $9\times9$ kernel and stride of 2. The volume is now sliced along its depth into 32 different layers of 8-dimension capsules, for a total of $6\times6\times32$ capsules. The third layer (DigitCaps) is a fully connected layer of 10 different 16-dimension capsules, with each capsule corresponding to an output class (a.k.a. 10 digit). Note that we have dynamic routing only between the PrimaryCaps and DigitCaps layers. Finally, we take the magnitude of these 16-dimension embeddings and output a final predicted  corresponding to the embedding with the largest magnitude.

The purpose of the second half of the model (Figure~\ref{reconsArch}), also referred to as the decoder, is to implement reconstruction as a regularization method. The 16-dimension embeddings from each class are concatenated, with all but the winning class's vector components masked to 0. Ten digit classes would mean a final embedding of length 160, which is then fed through 3 fully connected layers with 512, 1024, and 764 neurons respectively. The final 784-dimension  output is reshaped into a $28\times28$ reconstruction of the ground truth.

\subsection{Loss Function}

The two halves of the model are combined to train the network with a two part loss function. The first term penalizes false predictions, with predictions corresponding to the DigitCaps embeddings with the largest magnitude. The second term penalizes reconstruction error, or the difference in the ground truth image and its reconstruction after passing its DigitCaps embedding through the fully connected layers. We can use backpropagation to minimize the loss since dynamic routing is differentiable when unrolled.

\subsection{Our Contributions}

Based on the aforementioned sections, one cannot help but question the  design decisions made in making CapsNets work on MNIST. This is the root of our previously mentioned project goals, and now that we have properly explained CapsNets, we can talk about them in more detail.

First, we will explore whether this architecture will perform well on more difficult datasets. Unlike MNIST, other datasets may include images with noise, color, affine transformations, intra-class variation, natural scenes, and a variety of other factors that may or may not work well with the capsule architecture. We will also explore the effect of the number of routing iterations on performance, balanced against complexity constraints.

Second, to better understand the source of error and the expressiveness of the underlying model, we will try to understand and visualize what sorts of representations of the underlying data the model is able to capture. Our work adds to the existing literature by extending CapsNet to new datasets and providing unique visualizations that bring to light novel insights on CapsNets.

\section{Approach and Implementation}

\subsection{Datasets}

It has been demonstrated that CapsNets can achieve state of the art on MNIST. We want to try them on datasets that are marginally harder in specific ways, so that even if they fail we can gain insight into the nature of the limitations of CapsNets.

\begin{figure}[ht]
\centering
\begin{subfigure}{.5\textwidth}
	\centering
    \includegraphics[width=0.9\linewidth]{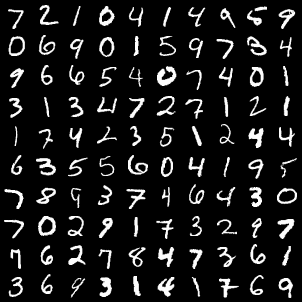}
    \caption{MNIST}
    \label{fig:dataset_mnist}
\end{subfigure}%
\begin{subfigure}{.5\textwidth}
	\centering
    \includegraphics[width=0.9\linewidth]{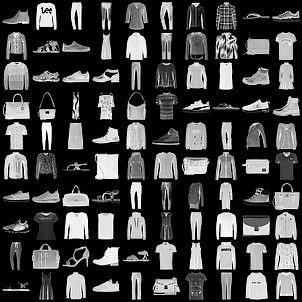}
    \caption{Fashion-MNIST}
    \label{fig:dataset_fashion}
\end{subfigure}
\begin{subfigure}{.5\textwidth}
	\centering
    \includegraphics[width=0.9\linewidth]{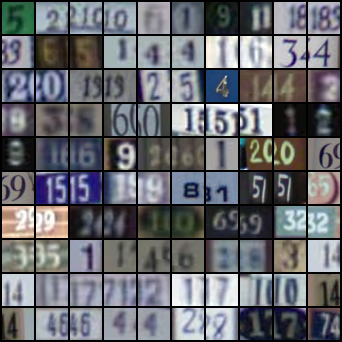}
    \caption{SVHN}
    \label{fig:dataset_svhn}
\end{subfigure}%
\begin{subfigure}{.5\textwidth}
	\centering
    \includegraphics[width=0.9\linewidth]{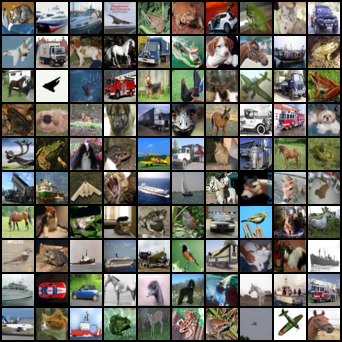}
    \caption{CIFAR10}
    \label{fig:dataset_cifar10}
\end{subfigure}
\caption{Dataset visualizations.}
\label{fig:dataset}
\end{figure}

\begin{enumerate}
\item \textbf{MNIST (\cite{lecun-mnisthandwrittendigit-2010}, Figure~\ref{fig:dataset_mnist})} The standard set of normalized and centered $28\times28$ black and white images of handwritten digits (0-9). 60,000 training samples and 10,000 testing samples.
\item \textbf{Fashion-MNIST (\cite{xiao2017fashion}, Figure~\ref{fig:dataset_fashion})} Exactly like MNIST except the image classes are items of clothing (from t-shirts to ankle boots).
\item \textbf{SVHN (\cite{netzer_reading_2011}, Figure~\ref{fig:dataset_svhn})} Contains cropped RBG $32\times32$ pixel images of house number digits taken from Google Street View. Like MNIST in that it is digits, but more complex because it has varying colors and styles, and multiple digits could be in a single sample. Is also larger, with 73257 training samples and 26032 testing samples.
\item \textbf{CIFAR-10 (\cite{krizhevsky_learning_2009}, Figure~\ref{fig:dataset_cifar10})}  Real-world objects database of RBG $32\times32$ pixel images across 10 classes, including vehicles (airplane, automobile, ship, and truck) and animals (bird, cat, deer, dog, frog, and horse). 50,000 training samples and 10,000 testing samples.

\end{enumerate}

Testing on MNIST will provide us an opportunity to verify the results of \cite{sabour_dynamic_2017}. Testing on Fashion-MNIST and CIFAR10 will allow us to see whether CapsNets generalize to different types of data. Testing SVHN and CIFAR10 again will allow us to see whether CapsNets generalize to color and to more complexity and intraclass variation.

\subsection{Deformations}

On top of these four datasets, we want to introduce some synthetic deformations to see how robust CapsNets are to new data. Specifically, we want to see how well they can classify data which is unlike that which they have seen before, but which still belongs to the same broader distribution. In other words, we want to see the extent to which CapsNets are deformation invariant for classification.

For each of the four datasets described above (MNIST, Fashion-MNIST, SVHN, CIFAR10) we generate an alternate, deformed dataset by applying a random affine deformation consisting of:
\begin{enumerate}
\item \textbf{Rotation}: Rotated image by a uniformly sampled angle within $[-20^{\circ},20^{\circ}]$.
\item \textbf{Shear}: Sheared along $x$ and $y$ axes by uniformly sampling shear parameters within $[-0.2,0.2]$. (Shear parameters are numbers added to the cross-terms in the $2\times3$ matrix describing an affine transformation.)
\item \textbf{Translation}: Translated along $x$ and $y$ axis by a uniformly sampled displacement parameters within $[-1,+1]$. (Displacement parameters are numbers added to the constant terms in the $2\times3$ matrix.)
\item \textbf{Scale}: Always scaled image 150\%.
\end{enumerate}

\begin{figure}
\centering
\begin{subfigure}{.5\textwidth}
	\centering
    \includegraphics[width=0.9\linewidth]{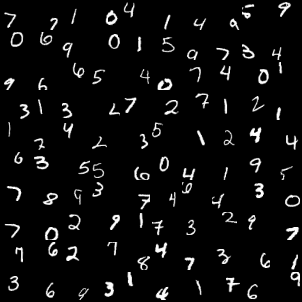}
    \caption{MNIST}
    \label{fig:dataset_mnist_trans}
\end{subfigure}%
\begin{subfigure}{.5\textwidth}
	\centering
    \includegraphics[width=0.9\linewidth]{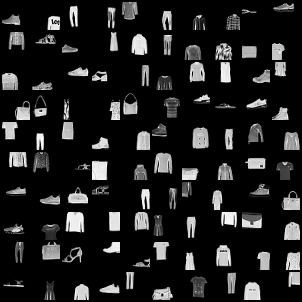}
    \caption{Fashion-MNIST}
    \label{fig:dataset_fashion_trans}
\end{subfigure}
\begin{subfigure}{.5\textwidth}
	\centering
    \includegraphics[width=0.9\linewidth]{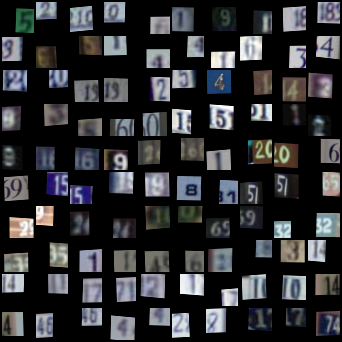}
    \caption{SVHN}
    \label{fig:dataset_svhn_trans}
\end{subfigure}%
\begin{subfigure}{.5\textwidth}
	\centering
    \includegraphics[width=0.9\linewidth]{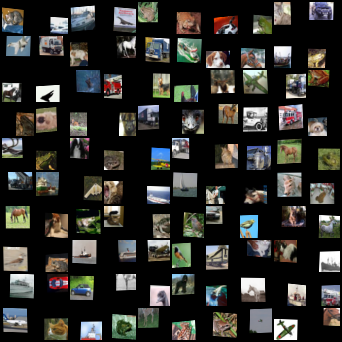}
    \caption{CIFAR10}
    \label{fig:dataset_cifar10_trans}
\end{subfigure}
\caption{Affine dataset visualizations.}
\label{fig:dataset_trans}
\end{figure}

In our experiments on robustness to transformation, we trained the CapsNet on the original versions with the datasets and tested on their affine-deformed versions. 100 examples of affinely transformed testing data are shown in Figure~\ref{fig:dataset_trans}.

\subsection{Baseline}

We need to baseline the performance of our ConvNet against a neural network architecture with a similar complexity (not only in number of parameters, but also, more importantly, in runtime). We chose to baseline against AlexNet, specifically a modification of the implementation put forth in the "One Weird Trick..." paper which  proposes the parallelization of the training of CNNs across GPUs (\cite{AlexNet}).  We retain the 5 traditional AlexNet layers of repeated convolution, ReLU non-linearities, and max-pooling, but we replace the final three fully connected hidden layers with a single fully connected layer of only 256 neurons. This produces performance that reaches close to state-of-the-art.

\subsection{Implementation}

\subsubsection{Code}
We started by forking a GitHub repository (\cite{iwasaki_capsule-networks:_2018}) of a PyTorch CapsNet implementation with an architecture matching the one used in the dynamic routing paper. This repository was the most starred implementation on Github. \textbf{Important edit in December 2020:} in June 2018 it was discovered that this implementation of CapsNets deviates from the specifications in the original paper (\cite{gram_ai_deviations}). We made some changes to enable the different ways we wanted to stretch CapsNets. Our code can be found at our open-sourced repository, \cite{keselj_capsule_2018}.

\begin{enumerate}
\item \textbf{Datasets} We enabled the use of additional datasets beyond MNIST, specifically Fashion-MNIST, SVHN, and CIFAR10, by adding new data loading and processing methods.
\item \textbf{Reconstructions} At the end of every epoch, we extract the DigitCapsule embeddings, reconstruct every class by running it through the decoder, and store all the reconstructions.
\item \textbf{Perturbations} Also at the end of every epoch, we perturb each element of the winning capsule embedding and store the reconstructions which result from them.
\item \textbf{Affine Transformations} We made a class which takes in a normal dataset and outputs the randomly affine transformed version of that dataset.
\item \textbf{AlexNet} We added basic support for training an AlexNet model under the same conditions as our CapsNets.
\item \textbf{Visualizations} The original code had basic error and accuracy logging on Visdom, but we moved it to TensorBoard and added displays of all our reconstructions, perturbations. The move to TensorBoard was primarily motivated by its useful dimensionality reduction suite, which allows us to visualize the internal embeddings of the model.
\end{enumerate}

\subsubsection{Computing}
Apart from these infrastructural changes, we had to run many trials under different parameters dozens of trials to train all of the CapsNets models using various hyperparameters and datasets. To quicken and parallelize the training process, we implemented our models with CUDA so that we could leverage GPUs to decrease training time. All three of the authors had access to GPUs associated with the labs in which they are conducting their theses. Prem and Rohan used 8 GPUs as part of the Visual AI Lab and Stefan has access to 6 GPUs associated with the Seung lab (all GPUs were GEFORCE GTX 1080s). We ran at least 30 CapsNet runs for our results in this paper, each taking about 4 hours, so a lower bound of our GPU-hours used is 120.

\section{Results and Discussion}

\textbf{Important edit in December 2020:} in June 2018 it was discovered that the implementation of CapsNets we used deviates from the specifications in the original paper (\cite{gram_ai_deviations}).

\subsection{Performance}
\subsubsection{Different Datasets}
We were able to apply CapsNets and AlexNet to our four dataset to achieve reasonable, but far from state of the art results. The training and testing curves for CapsNets and AlexNet are shown in below in Figure~\ref{train_test} and the end accuracies after 50 epochs are shown in the first half of Table~\ref{test_affine_vs_non_affine}.

We already know from the original paper that CapsNets can achieve almost state of the art MNIST, and here we verified that result by getting 99.5\% accuracy. CapsNets did relatively well on Fashion-MNIST and SVHN, with 89.8\% and 91.06\% accuracies, respectively. This makes sense because they are very similar to MNIST; Fashion-MNIST is in the same style and SVHN is of similar content. It is safe to say that CapsNets performed poorly on CIFAR10, achieving only 68.53\% accuracy before train and test error began diverging. This was expected, since the intra-class variation and background noise is more complex than that of MNIST .

CapsNets outperformed AlexNet in every case. For MNIST and Fashion-MNIST, the difference was marginal (1.03\% and 2.24\%, respectively), but in  SVHN and CIFAR10 it was more substantial (9.01\% and 7.92\%, respectively). When making this comparison, it is important to note that neither model is state-of-the-art, so it is not really an apples-to-apples comparison of architectures. Furthermore, we must emphasize that both models' hyperparameters are not optimal; for CapsNets we used the hyperparameters  used in \cite{sabour_dynamic_2017} (which was optimized for MNIST), and for AlexNet, we used PyTorch's default parameters (however, we decreased the learning rate for some datasets if the model could not converge). Nonetheless, this first-pass sanity check confirms that there is something interesting about CapsNets, since they can outperform CNNs that have many more parameters.

\begin{figure}
\centering
\includegraphics[width=\textwidth]{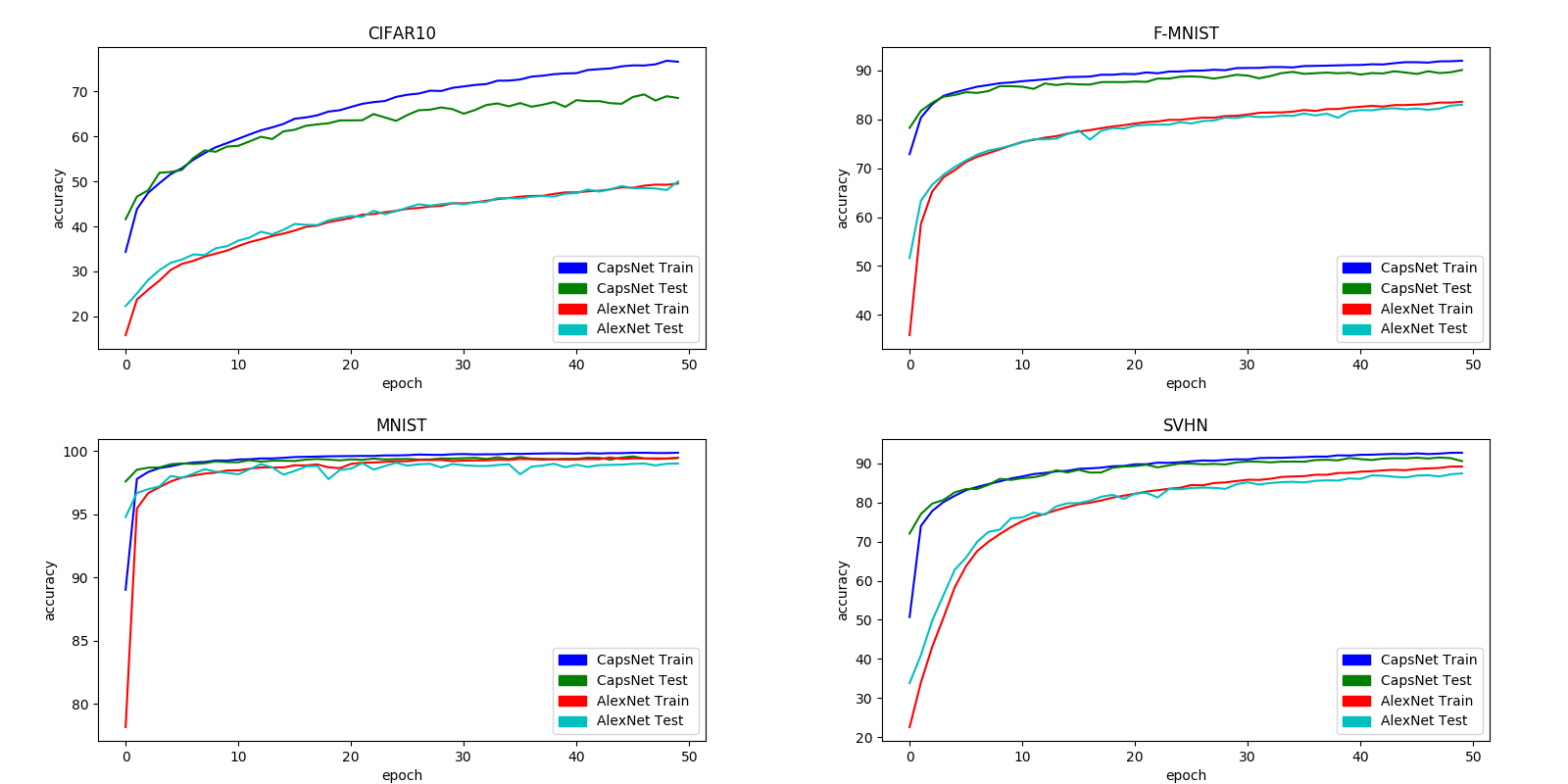}
\caption{CapsNet vs. AlexNet: Train and Test Curves for 50 epochs on 4 datasets (MNIST, Fashion-MNIST, SVHN, and CIFAR-10) without deformation.}
\label{train_test}
\end{figure}
 
\subsubsection{Affine Deformations} 
 
\begin{table}[ht]
\centering
\caption{CapsNet v. AlexNet: Classification test accuracy across 4 datasets after 50 epochs. With and without affine deformations.}
\label{test_affine_vs_non_affine}
\begin{tabular}{|c|cc|cc|c|}
\toprule
Dataset   & \multicolumn{2}{|c|}{Normal} & \multicolumn{2}{|c|}{Affine} \\
Network   & CapsNet & AlexNet & CapsNet  & AlexNet \\ \midrule
MNIST    & {\bf 99.50}    & 98.47    & 42.75   &   {\bf 55.21}  \\
Fashion-MNIST  & {\bf 89.80}    & 83.00    & {\bf 30.01}   &   25.80  \\
CIFAR10  & {\bf 68.53}    & 49.97    & {\bf 22.89}   &   20.21  \\
SVHN     & {\bf 91.06}    & 87.43    & {\bf 24.24}   &   22.86  \\
\bottomrule
\end{tabular}
\end{table}

A motivating reason for the study of CapsNets is the claim that they can recognize entities regardless of their configuration, unlike CNNs. As a test for this hypothesis, we trained CapsNet and AlexNet on the training sets from the four datasets above, but then tested on an affinely transformed version of their test sets, as described in our approach.

The final accuracies achieved by each model are shown in Table~\ref{test_affine_vs_non_affine}. Surprisingly, AlexNet outperforms CapsNet on deformed test data for MNIST, even though it had a lower accuracy on the normal test data. Still, CapsNets performed better on the affine-deformed versions of Fashion-MNIST, CIFAR-10, and SVHN datasets with more complex samples and greater intraclass variation. One explanation is that AlexNet has enough parameters to memorize the ten simple classes of MNIST, resulting in over-fitting. This is why its performance was not able to generalize to more complex classes. In that case, CapsNets may be learning a more generalizable representation of the data.

However, it could be the case that our AlexNet is simply a poor representative of CNNs, so it performing poorly on deformed data does not mean all CNNs would. To get a better idea about how capsule networks verus CNNs perform on this task in general, it might be interesting to consult a reasonable proxy for how poorly a model generalizes to deformed test data: the drop in accuracy when testing on the deformed data versus the normal data. In every dataset, CapsNet had a larger drop than AlexNet (56.75 \% vs 43.26\% for MNIST, 59.79 \% vs 57.20\% on Fashion-MNIST, 66.82\% vs 64.57 on SVHN, and 45.64\% vs 29.76\% on CIFAR10). This is raises doubts about whether CapsNets are actually better at capturing spacial relationships, because a model which builds spatially dependent representations should be better at generalizing to deformed data. It is perhaps the case that our AlexNet ``cheated'' by starting out with a poor performance and then not dropping that much. This is at least in part true (AlexNet had 49.97\% accuracy on normal CIFAR10), but it is not true in every case (AlexNet had 98.47\% accuracy on normal MNIST).

\subsubsection{Number of Routing Iterations}

It is evident that CapsNets need to improve their performance on datasets other than vanilla MNIST if they are to be as useful as CNNs. As mentioned above, we did not spend much time optimizing hyperparameters in our initial CapsNet vs. AlexNet tests, but now is a good time to dive into that. We found that varying the standard hyperparameters describing SGD (batch size, learning rate, learning rate decay, and momentum) did not change final performance much. This is  because our CapsNets were already converging well so these parameters could only decrease the training time.

The hyperparameter which most interests us is the number of routing iterations (from the dynamic routing algorithm). This parameter is unique to CapsNets and has important implications on their performance and runtime: every time inference is run, this many routing operations are run to determine which DigitCapsules the PrimaryCapsules send their information to. \cite{sabour_dynamic_2017} recommends 3 routing iterations, but we are curious about what the best number would be with respect to all four of our datasets.

We ran a set of experiments where we set the number of routing operations to one of (1, 2, 3, 4, 5) for each dataset. Although 3 iterations yielded good test accuracies, overall 2 performed about as well, even performing better in some instances. This is interesting for two reasons. First, it seems to go against the original paper's claim. Perhaps it could be the case that 3 iterations converges more stably than 2, but we reran our experiments multiple times and got similar graphs each time. Second, it could be indicative that the proposed dynamic routing algorithm is too ``abrupt''. The process of deciding which PrimaryCapsules feed into which DigitCapsules is a complex one, and would be surprising if an algorithm would only ever need two iterations to do this well.

\begin{figure}
\centering
\includegraphics[width=\textwidth]{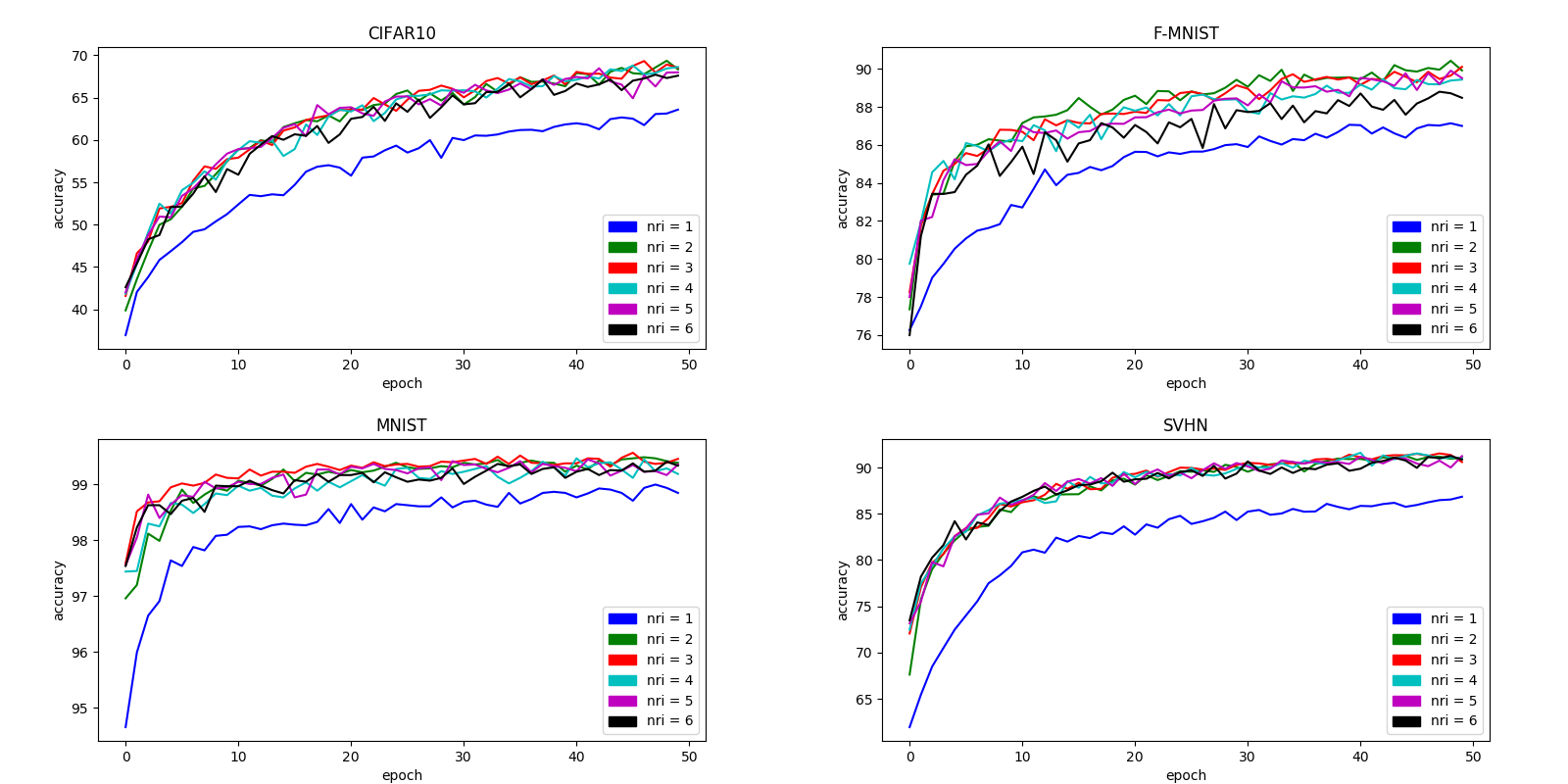}
\caption{}
\label{fig:effect_of_nri}
\end{figure}
\subsection{Embedding Spaces}

\subsubsection{Reconstructions}

\begin{figure}
\centering
\includegraphics[width=\textwidth]{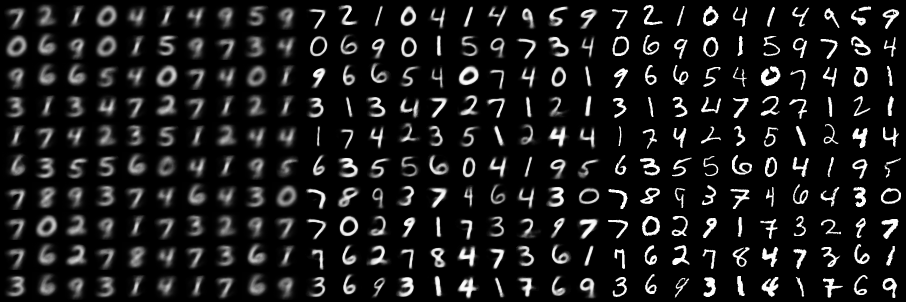}
\caption{Epoch 2 reconstruction, epoch 50 reconstruction, and ground truth for MNIST.}
\label{fig:reconstruction_mnist}
\end{figure}

\begin{figure}
\centering
\includegraphics[width=\textwidth]{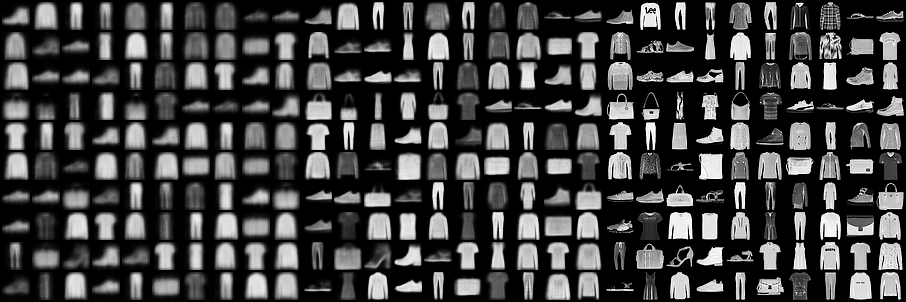}
\caption{Epoch 2 reconstruction, epoch 50 reconstruction, and ground truth for Fashion-MNIST.}
\label{fig:reconstruction_fashion}
\end{figure}

\begin{figure}
\centering
\includegraphics[width=\textwidth]{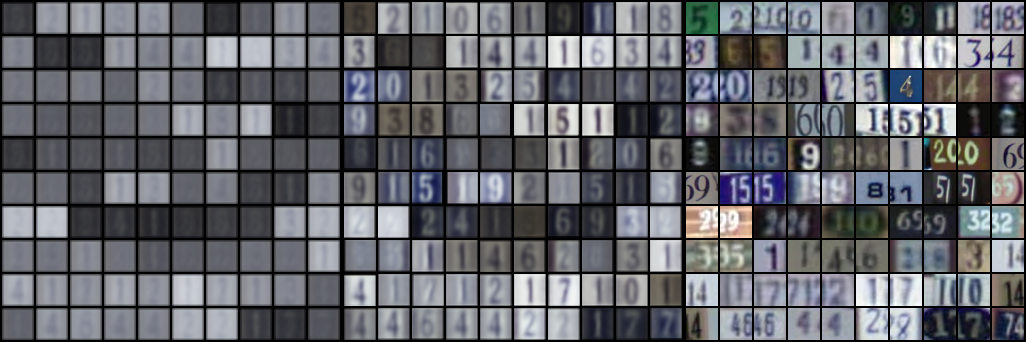}
\caption{Epoch 2 reconstruction, epoch 50 reconstruction, and ground truth for SVHN.}
\label{fig:reconstruction_svhn}
\end{figure}

\begin{figure}
\centering
\includegraphics[width=\textwidth]{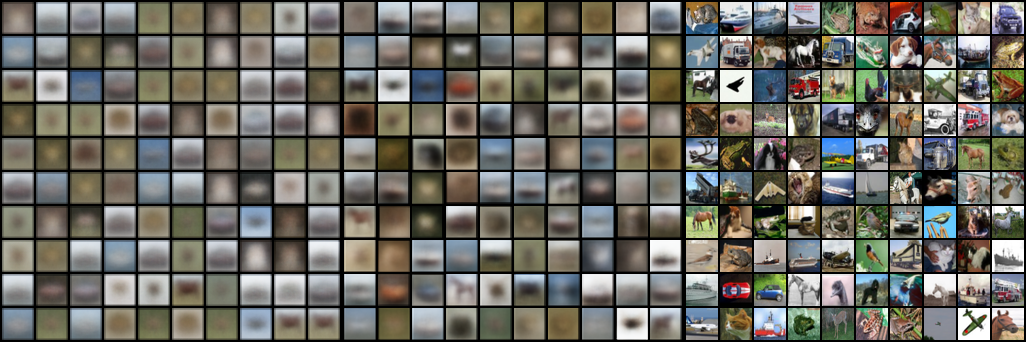}
\caption{Epoch 2 reconstruction, epoch 50 reconstruction, and ground truth for CIFAR10.}
\label{fig:reconstruction_cifar10}
\end{figure}

Now that we have an idea of the performance of CapsNet on various tasks, we want to understand why they are behaving the way they are so we can improve them. We can get a direct look into one component of the network's loss function by visualizing the reconstructions it generates. In Figures~\ref{fig:reconstruction_mnist},~\ref{fig:reconstruction_fashion},~\ref{fig:reconstruction_svhn}, and~\ref{fig:reconstruction_cifar10} we have shown the network's reconstruction after epoch 2, its reconstruction after epoch 50, and the ground truth images for each dataset.

MNIST's reconstruction improved substantially from epoch 2 to 50, mostly matching ground truth. However, the classification accuracy began to plateau earlier than the reconstruction accuracy, which suggests that the reconstruction does not necessarily help much with classification. Fashion-MNIST also had good reconstruction improvement, but the finer clothing details were not captured, which could be an effect of an embedding that is too small, or a decoder that is too shallow. SVHN went from failing to reconstruct anything substantial to a mostly gray rendering of the numbers. The network still failed to replicate colors. The reconstructions of CIFAR10, were always unrecognizable, apart from horses, which we assume is due to the consistency of point of view for most horse photos.

\subsubsection{Understanding Errors}

\begin{figure}[H]
\centering
\begin{subfigure}{.5\textwidth}
	\centering
    \includegraphics[width=0.8\linewidth]{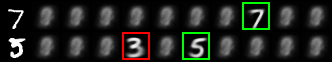}
\end{subfigure}%
\begin{subfigure}{.5\textwidth}
	\centering
    \includegraphics[width=0.8\linewidth]{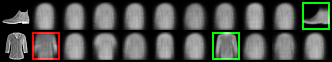}
\end{subfigure}
\begin{subfigure}{.5\textwidth}
	\centering
    \includegraphics[width=0.8\linewidth]{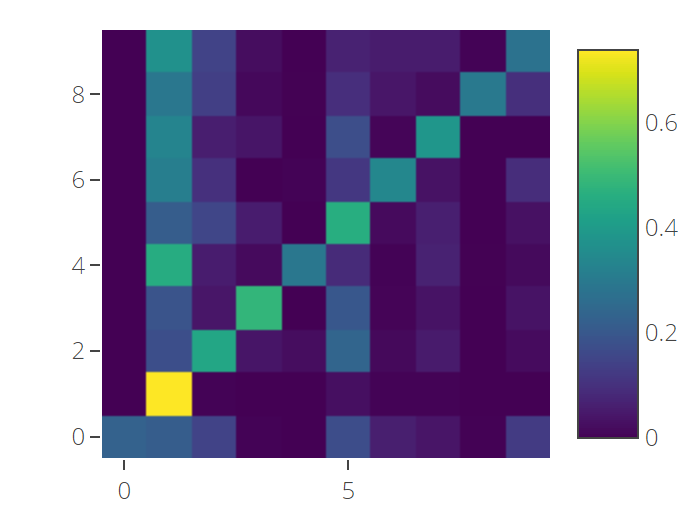}
    \caption{MNIST (deformed)}
    \label{fig:confusion_mnist}
\end{subfigure}%
\begin{subfigure}{.5\textwidth}
	\centering
    \includegraphics[width=0.8\linewidth]{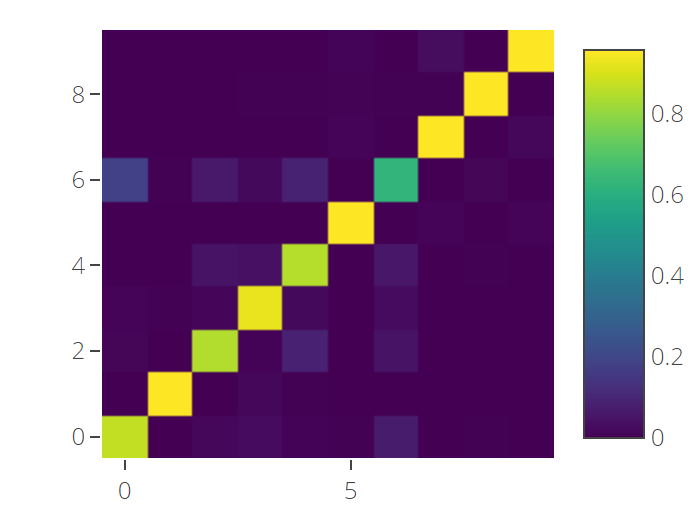}
    \caption{Fashion-MNIST}
    \label{fig:confusion_fashion}
\end{subfigure}
\caption{Confusion matrices and example correct and incorrect classifications for MNIST (affine deformation) and Fashion-MNIST.}
\label{fig:confusion}
\end{figure}

For the datasets with the best reconstructions, MNIST and Fashion-MNIST, we can make use of their reconstructions when understanding error sources, since we can decode the embeddings and recover visualizations of the misclassification the network made, as seen in Figure~\ref{fig:confusion}. Normally, the decoder accepts the 160-length embedding vector masked to 0 in all but the section that corresponded to the most confident class, and uses it to reconstruct the image. By simply unmasking each class's section in turn, we can generate visualizations for each class.

When the network is very confident and gets something right, as seen in the top row of reconstructions in Figure~\ref{fig:confusion}, the other class's components are small enough such that even without masking we don't see any meaningful reconstructions. However, as seen in the bottom row of reconstructions, when you reconstruct the correct class of most errors, a meaningful result is observed. The errors are genuinely ambiguous from a human perspective as well. In the MNIST case, the digit is somewhere between 5 and 3, and in the Fashion-MNIST case, the erroneous class is ``T-shirt/top'' while the correct class is ``Shirt''.

\subsubsection{Perturbations}
\begin{figure}
\centering
\includegraphics[height=6cm]{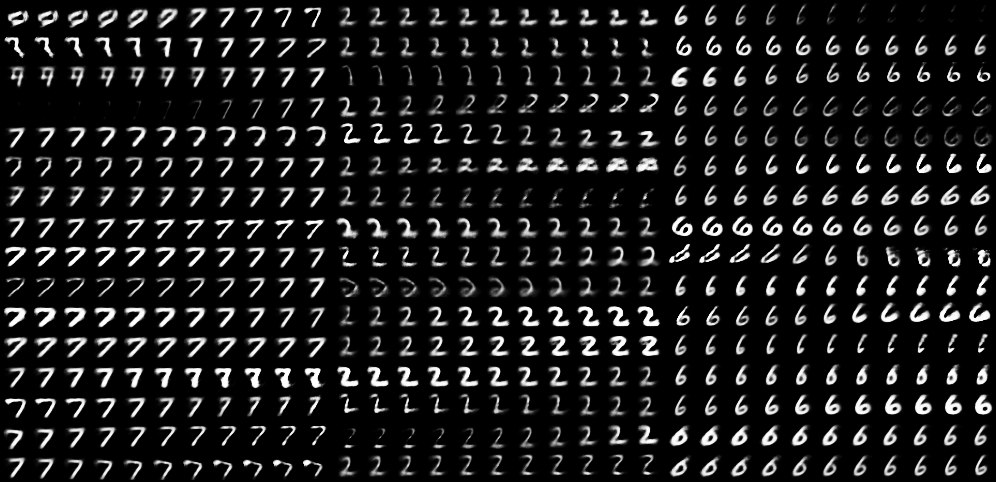}
\caption{Perturbation of an MNIST sample embedding vector at epoch 34, 38, and 50 respectively.}
\label{fig:perturbation_mnist}
\end{figure}

\begin{figure}
\centering
\includegraphics[height=6cm]{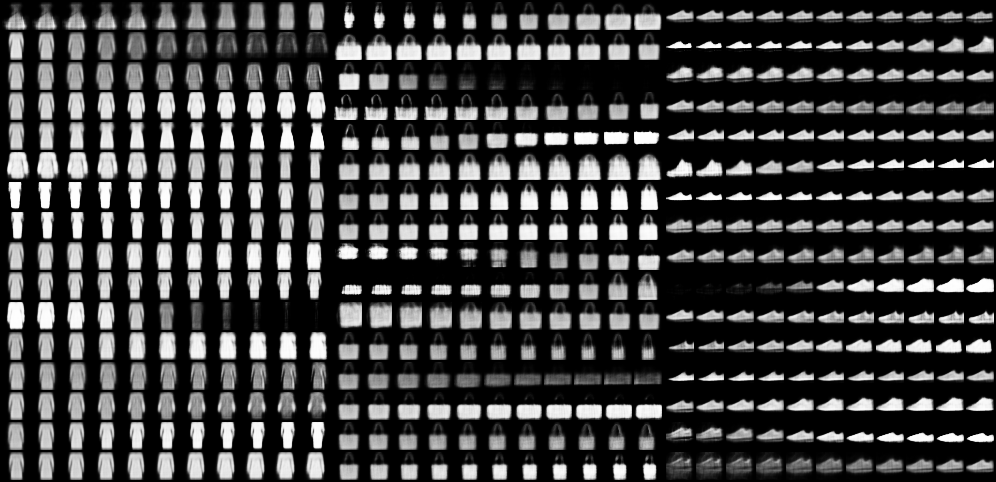}
\caption{Perturbation of a Fashion-MNIST sample embedding vector at epoch 33, 34, and 38 respectively.}
\label{fig:perturbation_fashion}
\end{figure}

\begin{figure}
\centering
\includegraphics[height=6cm]{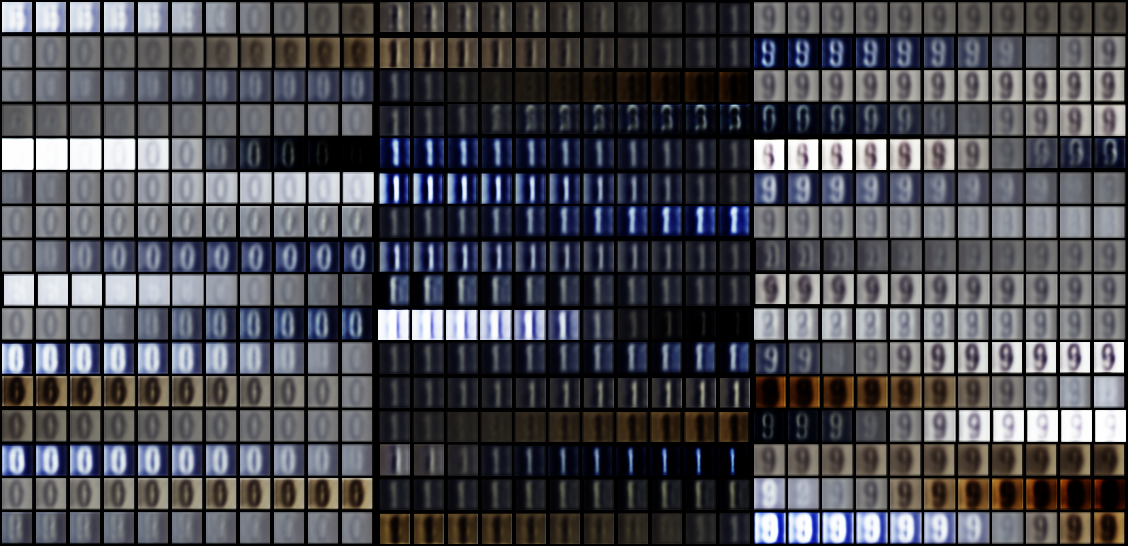}
\caption{Perturbation of an SVHN sample embedding vector at epoch 34, 38, and 50 respectively.}
\label{fig:perturbation_svhn}
\end{figure}

\begin{figure}
\centering
\includegraphics[height=6cm]{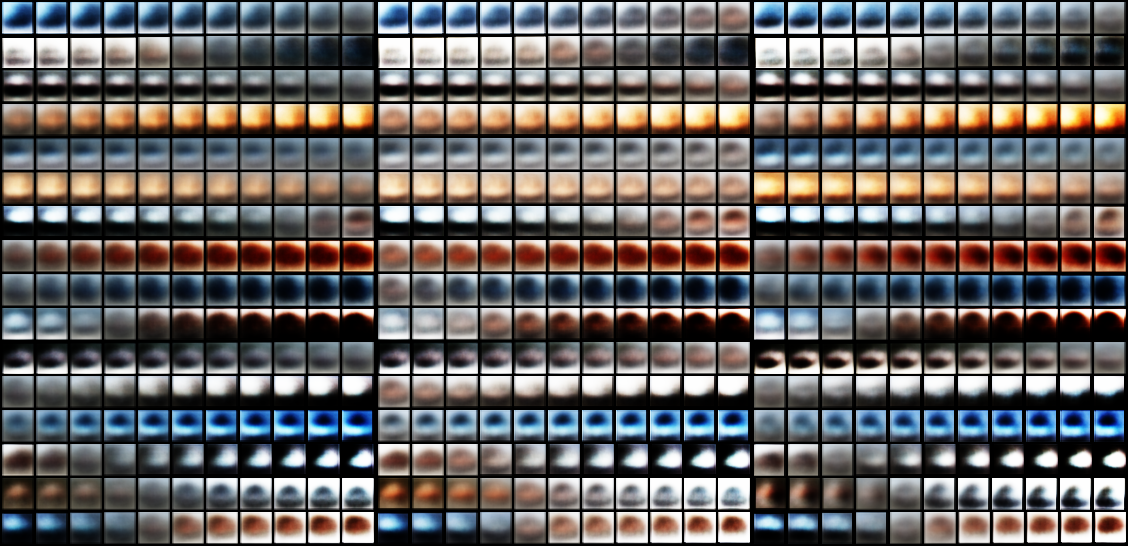}
\caption{Perturbation of a CIFAR10 sample embedding vector at epoch 34, 38, and 50 respectively.}
\label{fig:perturbation_cifar10}
\end{figure}

Similar to Figure~4 of \cite{sabour_dynamic_2017}, we want to qualitatively observe and understand the learned embedding space and see what is parameterized by each component for each class. The embedding vector for our model is 160 components long, 16 for each output class. All of it is masked away except for the section for the predicted output class. By replacing each of the 16 non-zero components of a sample input image's embedding vector with a scalar between~$-0.25$ and~$0.25$, we can observe what the high level meaning of these components is by running this perturbed vector through the network decoder, as seen in MNIST (Figure~\ref{fig:perturbation_mnist}), Fashion-MNIST (Figure~\ref{fig:perturbation_fashion}), SVHN (Figure~\ref{fig:perturbation_svhn}), and CIFAR10 (Figure~\ref{fig:perturbation_cifar10}). Each row is a different component, and each column is a substituted value of a multiple of~$0.05$ from~$-0.25$ to~$0.25$.

In MNIST, we can clearly see general components that correspond to stroke thickness, height, width, and skew, across all output digit classes. Then there are digit-specific ones that capture unique properties, like the whirl near the bottom left of some~2s, or the squashing of the enclosed part of the~6. In Fashion-MNIST, the embedding captures some general properties like height, width, grayscale color, and texture, and some specific properties like bag handle arc size and shape, the prominence of the tongue and topline for the shoe, or the arm length and waist size of the dress. SVHN components appear to mostly capture the color variation of the digits. Perturbations of CIFAR10 embeddings result in strong color changes as well. We suspect due to the shallowness of the decoder part of the network, both color and high class variance datasets have much worse reconstructions, and require a deeper network with larger embeddings.

\subsubsection{PCA Visualization of Dynamic Routing}

\begin{figure}
\centering
\includegraphics[width=0.9\textwidth]{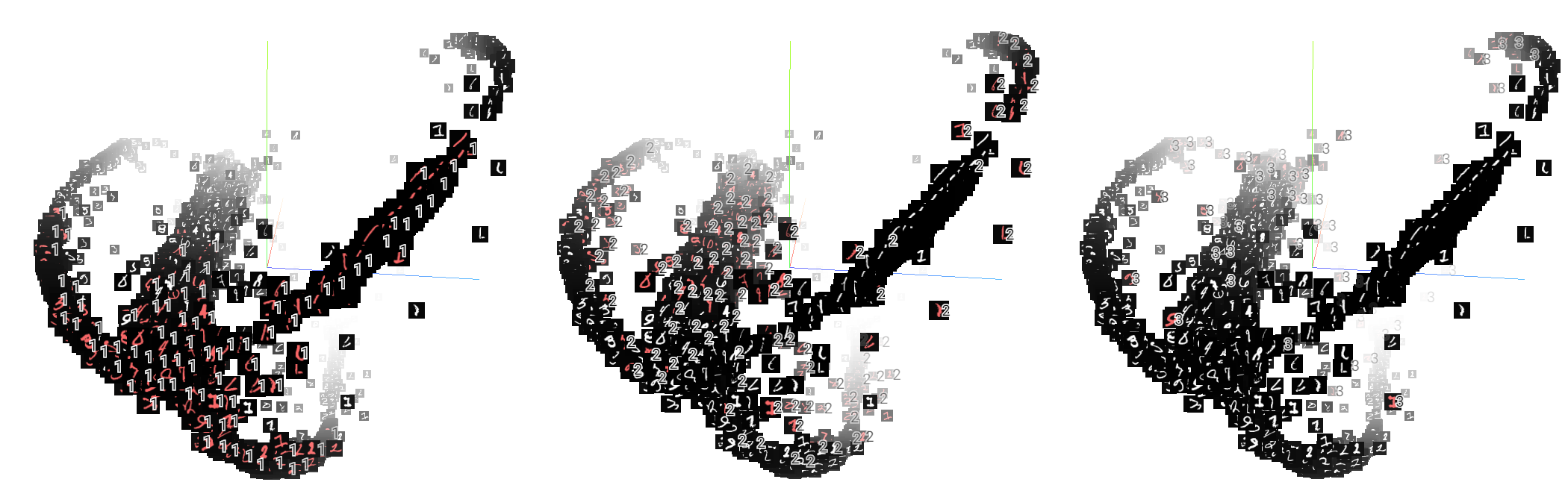}
\caption{3D sphereized PCA of drifting embedding vector outputs over 3 dynamic routing iterations (epoch 50 of MNIST).}
\label{fig:pca_mnist}
\end{figure}

\begin{figure}
\centering
\includegraphics[width=0.9\textwidth]{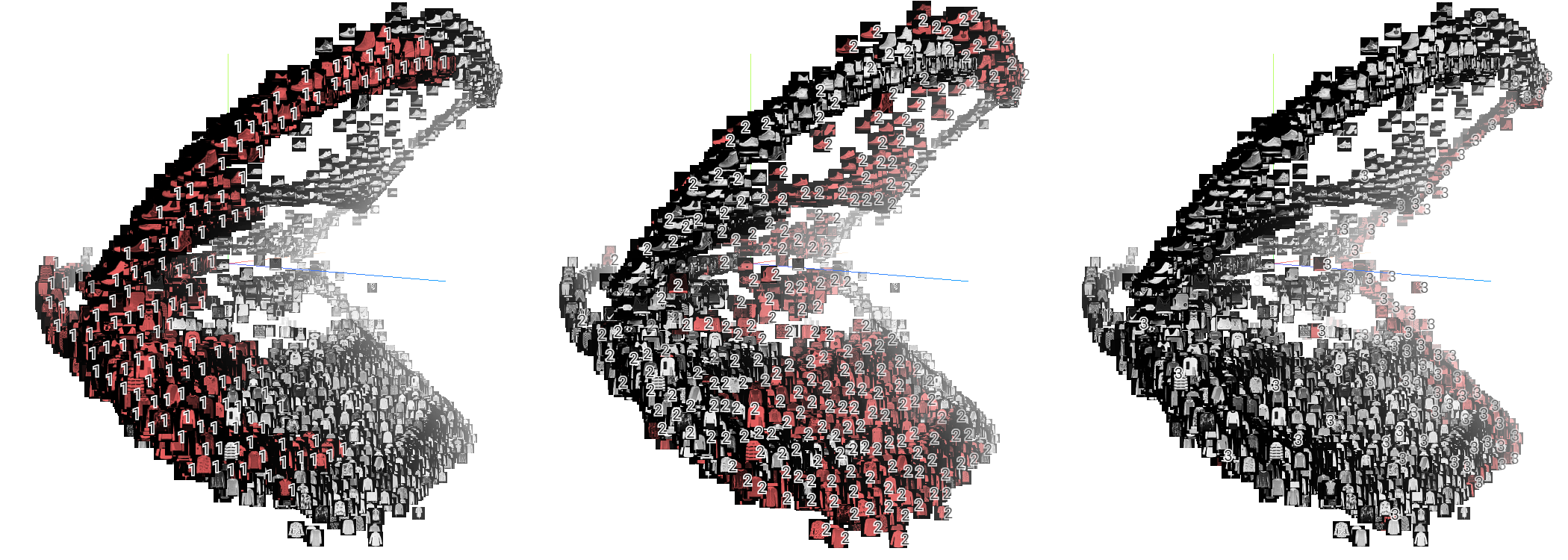}
\caption{3D sphereized PCA of drifting embedding vector outputs over 3 dynamic routing iterations (epoch 50 of Fashion-MNIST).}
\label{fig:pca_fashion}
\end{figure}

\begin{figure}
\centering
\includegraphics[width=0.9\textwidth]{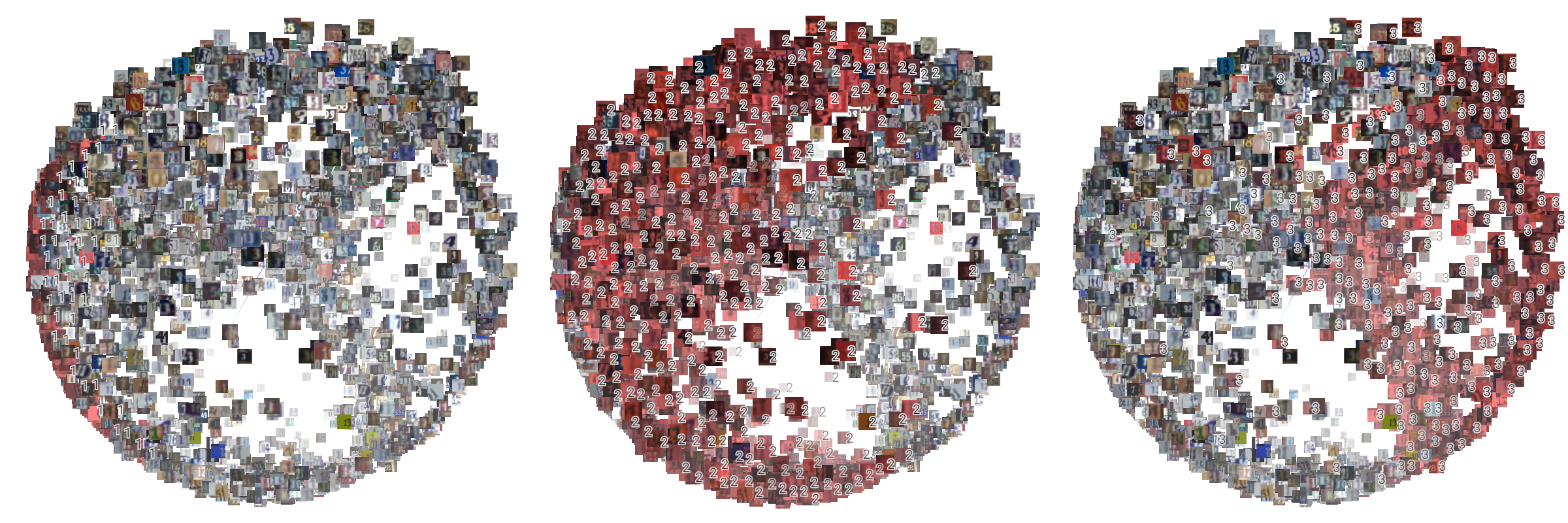}
\caption{3D sphereized PCA of drifting embedding vector outputs over 3 dynamic routing iterations (epoch 50 of SVHN).}
\label{fig:pca_svhn}
\end{figure}

We used TensorBoard to visualize the drift of the digit output embeddings over the 3 dynamic routing iterations during the forward pass (Figures~\ref{fig:pca_mnist}, \ref{fig:pca_fashion}, \ref{fig:pca_svhn}). In each figure, we can consistently see a large shift in the distribution of vectors as dynamic routing iteration progresses. The large spread of vectors generally in the second iteration gets reduced by the third, as they converge to one of the final classes. Despite the spread from the second iteration, we showed above that 2 was sufficient for achieving the accuracy from the recommendation of 3 in \cite{sabour_dynamic_2017}.

\begin{figure}
\centering
\begin{subfigure}{.5\textwidth}
	\centering
    \includegraphics[height=0.8\linewidth]{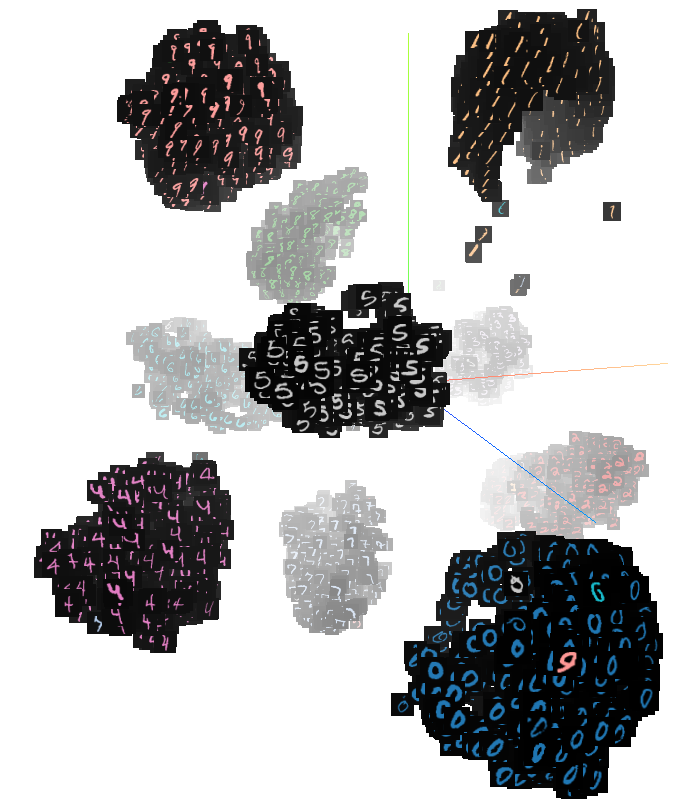}
    \caption{MNIST}
    \label{fig:tsne_mnist}
\end{subfigure}%
\begin{subfigure}{.5\textwidth}
	\centering
    \includegraphics[height=0.8\linewidth]{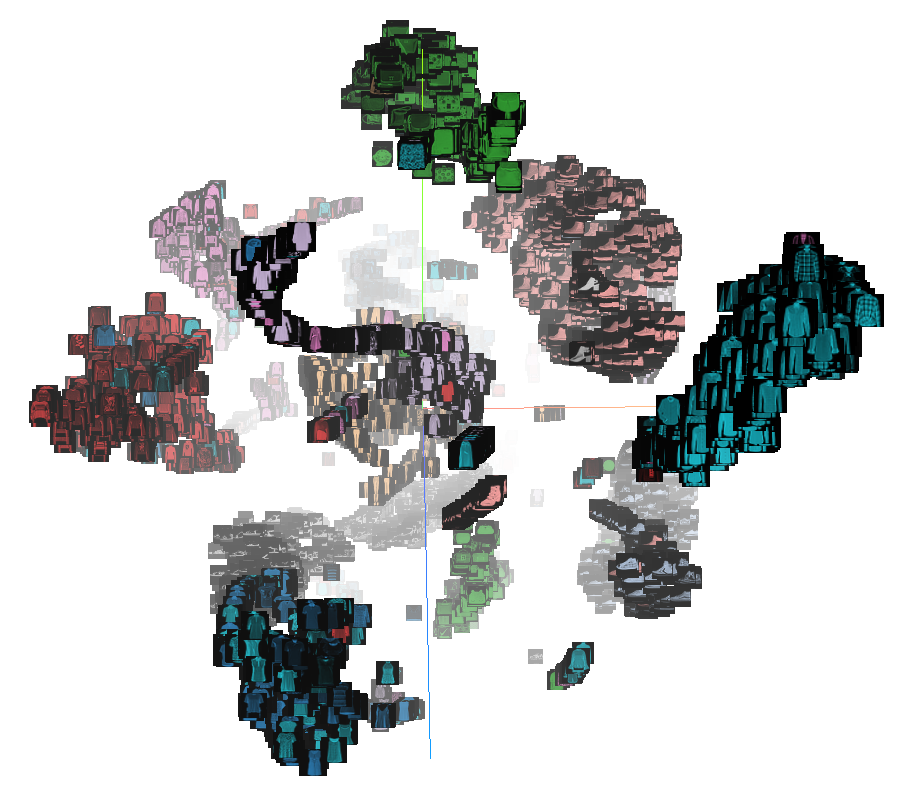}
    \caption{Fashion-MNIST}
    \label{fig:tsne_fashion}
\end{subfigure}
\begin{subfigure}{.5\textwidth}
	\centering
    \includegraphics[height=0.8\linewidth]{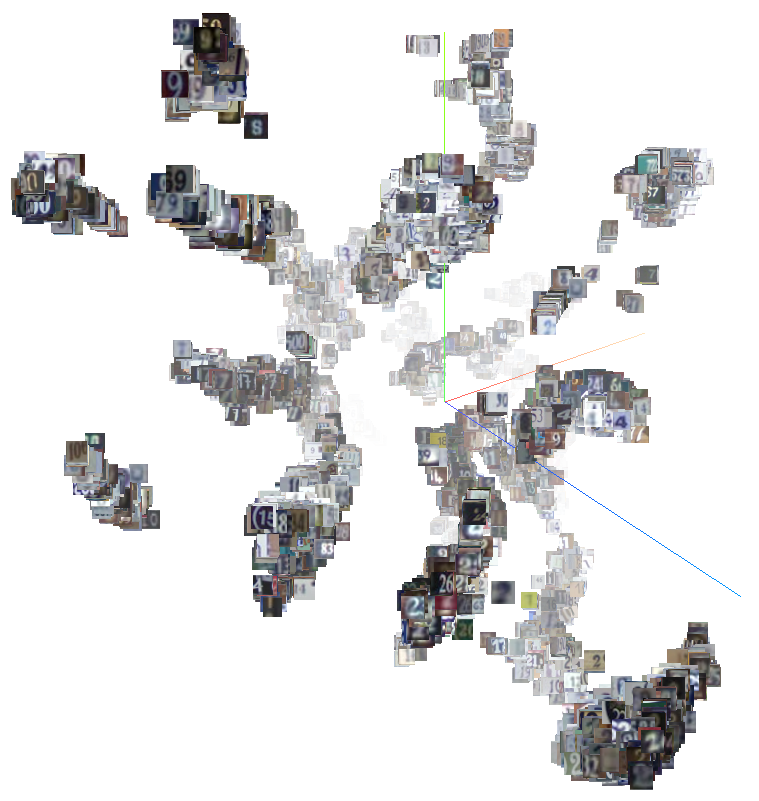}
    \caption{SVHN}
    \label{fig:tsne_svhn}
\end{subfigure}%
\begin{subfigure}{.5\textwidth}
	\centering
    \includegraphics[height=0.8\linewidth]{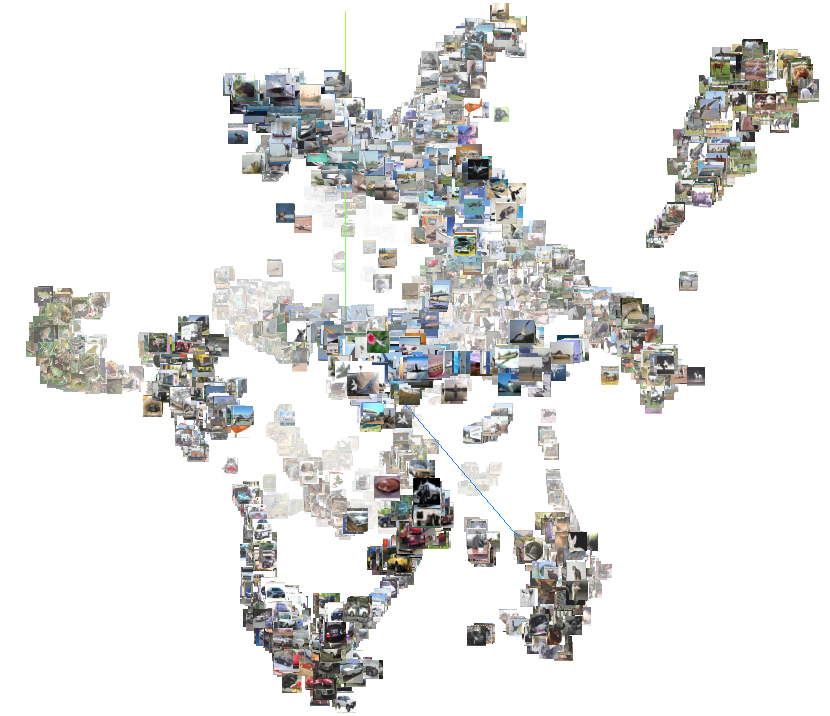}
    \caption{CIFAR10}
    \label{fig:tsne_cifar10} 
\end{subfigure}
\caption{t-SNE visualizations of 10k embeddings (50 epochs, 3rd dynamic routing iteration).}
\label{fig:tsne}
\end{figure}

\subsubsection{Applying t-SNE to Embeddings}

t-SNE visualizations project embeddings into a 2D or 3D space, putting similar vectors closer to each other in Euclidean distance. When coupled with the ground truth class labels, they visualize how close or far embeddings of the same class are from each other. We visualized all datasets in Figure~\ref{fig:tsne}.

Figure~\ref{fig:tsne_mnist} shows near perfect clustering into 10 groups for MNIST. However, each of the other datasets is not nearly so neat. Fashion-MNIST is next in complexity, and the worst is CIFAR10. The disorder of embedding appears correlated to the test accuracy and quality of reconstruction. This means that the t-SNE visualization is a potentially informative complement to accuracy for assessing strategies for improving network performance on more complex datasets.

\section{Conclusion and Future Work}

In this work, we pushed the limits of CapsNets by exploring their performance and expressiveness when controlled, incremental changes were made to them. Our experiments yielded findings that are pertinent to the goals we outlined at the beginning:
\begin{enumerate}
\item CapsNets were able perform better than AlexNet on datasets marginally harder than MNIST, but they were not as good at coping with deformations as one might expect. Increasing the number of routing iterations from the recommended 3 did not help, and in fact, decreasing it to 2 did not harm performance. This suggests that although CapsNets are probably learning some useful information about spatial relationships between entities, they are not making full use of routing to encode them robustly. 
\item CapsNets could convincingly reconstruct MNIST and Fashion-MNIST samples, whose embeddings seemed to measure meaningful qualities like thickness or skew. However, they could not do so for SVHN or CIFAR10, whose embeddings mainly encoded intensity and color. Across all datasets, most of the errors made by CapsNets corresponded to noisy or ambiguous reconstructions. The t-SNE visualizations of the embeddings further corroborated this, by showing that only MNIST had a cleanly separable label space.

\end{enumerate}

Together, our findings indicate that the current CapsNet design is unlikely to work on other classification tasks, let alone machine learning tasks in general. This being said, we are not recommending that the idea be discarded; the concept of a capsule is intuitively appealing and they have demonstrated reasonable performance in our experiments. We believe that there is significant potential for CapsNets to be improved and made useful. The three paths we are most excited about are: 

\begin{enumerate}
\item \textbf{Routing}
The current routing algorithm is perhaps the most simple and intuitive way of deciding which high level capsules get assigned to which low level capsules. Fleshing it out to be more informed and less brittle could enable CapsNets to learn more complicated structure in the data. For example, a neural network could act as the mechanism that reweighs coefficients. 
\item \textbf{Architecture}
The current architecture is also relatively austere. There is only one layer of convolution for feature generation and only one layer of capsules before the end object capsules are built. Although this shallow structure might work on MNIST, it seems unreasonable to expect that all the nuances of a vehicle or animal in CIFAR could be represented with one level of entities. A deeper network, perhaps with some domain specific structure, might overcome this. \cite{sabour_dynamic_2017} has some other experiments in this direction.
\item \textbf{Task}
Finally, it could be the case that CapsNets would be good at other more complicated tasks, even though they are not the best at the ``simple'' task of classification. The task we are most curious about is segmentation, because perhaps the cosine similarities between embeddings computed by the CapsNet could be used as indicators of how likely it is that two pixels are in the same object.
\end{enumerate}

\section{Acknowledgements}

We would like to thank Dr. Russakovsky and Dr. Ferencz for their tireless patience in teaching us the fundamentals of computer vision. This work would not be possible without the computational resources of the Visual AI Lab and Seung Lab and our starter code from Kenta Iwasaki of Gram.AI.
\newpage

\bibliographystyle{plainnat}
\bibliography{nips}

% \appendix
% \renewcommand\thefigure{\thesection.\arabic{figure}}   
% \setcounter{figure}{0}  
% \section{How many routing iterations to use?}

\end{document}